\documentclass[10pt,twocolumn,letterpaper]{article}

\usepackage[algorithms]{wacv} 
\usepackage{svg}
\usepackage{arydshln} 

\usepackage{float}
\usepackage{latexsym}
\usepackage{amssymb}
\usepackage{amsmath}
\usepackage{amsthm}
\usepackage{booktabs}
\usepackage{enumitem}
\usepackage{graphicx}
\usepackage{makecell}
\usepackage{url}
\usepackage{multirow}
\usepackage{color}
\usepackage[table]{xcolor}
\usepackage{threeparttable}
\usepackage{tabularray}
\usepackage{rotating}
\usepackage{tabularx}
\usepackage{placeins}
\usepackage{svg}
\usepackage{subcaption}
\usepackage[linesnumbered,ruled,vlined]{algorithm2e}

\usepackage{pifont} 

\definecolor{wacvblue}{rgb}{0.21,0.49,0.74}
\usepackage[pagebackref,breaklinks,colorlinks,allcolors=wacvblue]{hyperref}

\def\wacvPaperID{1424} 
\def\confName{WACV}
\def\confYear{2026}

\title{WWE-UIE: A Wavelet \& White Balance Efficient\\ Network for Underwater Image Enhancement}

\author{Ching-Heng Cheng \quad Jen-Wei Lee \quad Chia-Ming Lee \quad Chih-Chung Hsu\\
National Cheng Kung University \quad
National Yang Ming Chiao Tung University}

\begin{document}
\maketitle

\begin{abstract}
Underwater Image Enhancement (UIE) aims to restore visibility and correct color distortions caused by wavelength-dependent absorption and scattering. Recent hybrid approaches, which couple domain priors with modern deep neural architectures, have achieved strong performance but incur high computational cost, limiting their practicality in real-time scenarios. In this work, we propose WWE-UIE, a compact and efficient enhancement network that integrates three interpretable priors. First, adaptive white balance alleviates the strong wavelength-dependent color attenuation, particularly the dominance of blue-green tones. Second, a wavelet-based enhancement block (WEB) performs multi-band decomposition, enabling the network to capture both global structures and fine textures, which are critical for underwater restoration. Third, a gradient-aware module (SGFB) leverages Sobel operators with learnable gating to explicitly preserve edge structures degraded by scattering. Extensive experiments on benchmark datasets demonstrate that WWE-UIE achieves competitive restoration quality with substantially fewer parameters and FLOPs, enabling real-time inference on resource-limited platforms. Ablation studies and visualizations further validate the contribution of each component. The source code is available at \url{https://github.com/chingheng0808/WWE-UIE}.
\end{abstract}

\section{Introduction}

Underwater imaging supports critical applications in marine ecology, robotics, and infrastructure inspection~\cite{underwater-app}. However, light attenuation and scattering in water severely reduce contrast and distort colors, with rapid red absorption producing dominant blue-green tones~\cite{attenuation,scattering,uie-imgprocessing}. Underwater image enhancement (UIE) aims to restore visual fidelity and narrow the domain gap to terrestrial images, enabling reliable deployment in tasks such as autonomous navigation and real-time marine inspection~\cite{uie-absorption,domaingap, robotic,eich2014robot}.

\begin{figure}[H] 
\centering
\includegraphics[width=1.0\linewidth]{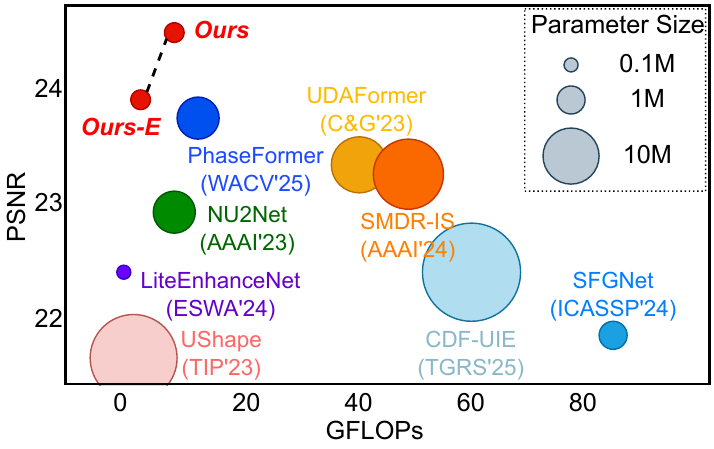}
\caption{Performance comparison between our method and its variant, and other state-of-the-art UIE methods on UIEB dataset~\cite{waternet}. FLOPs is measured through $256{\times}256$ input image.}
\label{fig:cp-plot}
\end{figure}

Early attempts in UIE were dominated by physics- and prior-driven methods~\cite{cong2024comprehensivesurveyunderwaterimage, conventional}, which rely on handcrafted assumptions such as transmission estimation or reflectance decomposition \cite{retinex-uie}. 
While physically interpretable, these approaches may be less reliable across diverse underwater conditions, where minor estimation errors can compound across stages \cite{cong2024comprehensivesurveyunderwaterimage}.
Convolutional Neural Networks (CNN)-based approach~\cite{liteenhancenet,atrous-cnn} usually operate with limited receptive fields, which restricts performance under severe scattering and color attenuation.
On the other hand, Transformer-based frameworks~\cite{udaformer,smdris} were introduced to capture long-range dependencies as well as achieve stronger global context modeling. However, it is challenging to achieve real-time deployment due to the higher computational complexity.
More recently, hybrid strategies such as WF-Diff \cite{wf-diff} and TCTL-Net \cite{TCTL} combined frequency cues or template-free priors with deep backbones. Although they improve global modeling and restoration quality, they either incur substantial computational cost or neglect explicit frequency decomposition, resulting in longer inference time and less reliable fine-structure recovery.

To address the trade-off between enhancement quality and efficiency in existing methods, we propose \textbf{WWE-UIE}, a compact and efficient underwater image enhancement (UIE) model designed for real-time scenarios. Rather than relying on computationally expensive transformer-based architectures, WWE-UIE enhances the performance of lightweight convolutional networks through a principled integration of interpretable domain priors. Specifically, white balance correction alleviates wavelength-dependent color attenuation, wavelet-based decomposition captures multi-band details utilizing efficient Haar wavelet, and a gradient-aware module preserves edge sharpness and fine details affected by scattering. In addition, the HVI color space~\cite{hvi} is incorporated into the loss function to guide the decoupling of luminance and chromaticity, facilitating better adaptation to underwater color variations. This synergistic framework, built upon spatial priors, multi-band cues, and perceptually motivated color representation, allows WWE-UIE to deliver robust enhancement results with minimal computational cost.
As shown in Fig.~\ref{fig:cp-plot}, our method achieves a favorable trade-off among PSNR, model size, and FLOPs.
In summary, the main contributions of this work can be listed as follows:
\begin{itemize}
    \item We present \textbf{WWE-UIE}, which judiciously combines our \textbf{adaptive white balance}, wavelet-based band decomposition, and gradient-aware refinement module to improve performance without increasing computational costs.  
    \item The Wavelet-based Enhancement Block (WEB) efficiently captures multiple bands to enhance both global color consistency and local details, and the Sobel Gradient Fusion Block (SGFB), which emphasizes edge sharpness and preserves structural fidelity under scattering. Both modules incur negligible computational cost.  
    \item We conducted extensive experiments on multiple benchmark datasets, showing that WWE-UIE achieves competitive performance with substantially fewer parameters and FLOPs, making it suitable for real-time deployment.  
\end{itemize}

\section{Related Works}

\subsection{Traditional UIE Methods}

\textbf{Physics-based methods} attempt to restore scene radiance by explicitly modeling wavelength-dependent attenuation and particle scattering. Early adaptations of the dark channel prior~\cite{he2010single, li2016underwater} compensate for haze-like effects with additional color correction, while Sea-Thru~\cite{Sea-thru} leverages range-dependent attenuation models for more faithful color recovery. Other designs, such as fuzzy-logic based HFM~\cite{hfm}, introduce flexible formulations to handle scattering. These approaches improve visibility and naturalness, yet their reliance on accurate depth, water type, or illumination parameters makes them fragile and hard to generalize across diverse underwater conditions~\cite{berman2020underwater}.  

\textbf{Model-free methods} enhance visibility using image statistics without relying on physical assumptions. Retinex formulations~\cite{retinex-uie} separate illumination and reflectance to boost local contrast. UMSHE \cite{hist-uie} introduces histogram equalization UIE method, which is based on partitioned local region statistics, to address global nonuniform drifting. \textit{Ancuti et al.}~\cite{cc7} fuses white-balanced and original input to compensate for the dominance of blue-green channels, as well as associated with the weight maps, which are defined to promote the transfer of edges and color contrast to the output image. WWPF~\cite{wavelet-fuse} further combines decomposed wavelet signals to refine clarity. 

While computationally simple, traditional UIE methods often suffer from oversaturation or artifacts under complex degradations, highlighting the value of domain priors but also the limitations of handcrafted assumptions in real-world scenarios.

\subsection{Learning-based UIE Methods}
\textbf{Spatial–Frequency Learning Methods.} Recent methods increasingly leverage the complementary strengths of spatial and frequency domains for underwater image enhancement, such as CDF-UIE \cite{cdfuie} and UShape \cite{ushape}. Spatial-domain approaches like SMDR-IS~\cite{smdris} and UDAFormer \cite{udaformer} integrate multiscale CNNs and attention mechanisms, capturing local and global structural information but often struggle with severe color attenuation or texture degradation~\cite{liteenhancenet,atrous-cnn}. Frequency-domain designs provide a natural remedy: models such as SFGNet \cite{sfgnet} and PhaseFormer \cite{phaseformer} employ Fourier transforms to learn in the frequency domain. This strategy enables targeted correction of low-frequency color shifts and high-frequency detail loss, improving tone fidelity. However, the global nature of Fourier-based operations often increases computational and memory costs.

\textbf{Color Transfer Methods.} Color transfer strategies address wavelength-dependent attenuation by rebalancing the color distribution of underwater images. Representative approaches include TCTL-Net \cite{TCTL}, which performs template-free color transfer in the CIELab space, and UColor \cite{ucolor}, which incorporates reverse medium transmission estimated by Generalization of the Dark Channel Prior (GDCP) to guide chromatic adjustment \cite{he2010single}. CECF \cite{CECF} learns the mapping from a distorted image to an enhanced image by the decomposition of color codes. While effective in alleviating color casts, these methods often rely on strong assumptions about color distributions or handcrafted priors, which may be fragile under diverse underwater conditions.

\textbf{Hybrid Methods.}
Hybrid methods integrate physical priors, frequency cues, and learnable architectures to improve robustness. ReX-Net \cite{Rex-Net} introduces Retinex-based reflectance estimation with CNN and attention modules, while SGUIE-Net \cite{SGUIE} leverages semantic priors to recover uncommon degradations. MFEF \cite{MFEF} integrates a white balance algorithm and the contrast-limited adaptive histogram equalization (CLAHE) algorithm to extract different forms of rich features in multiple views. WF-Diff \cite{wf-diff} integrates the Fourier-domain prior into a diffusion-based framework. 
Although hybrid methods effectively leverage domain knowledge, they often rely on complex or heavy architectures.
This motivates the development of lightweight prior-guided frameworks that balance interpretability, restoration quality, and efficiency.

\section{Methodology}

\begin{figure*}
    \centering
    \includegraphics[width=0.90\linewidth]{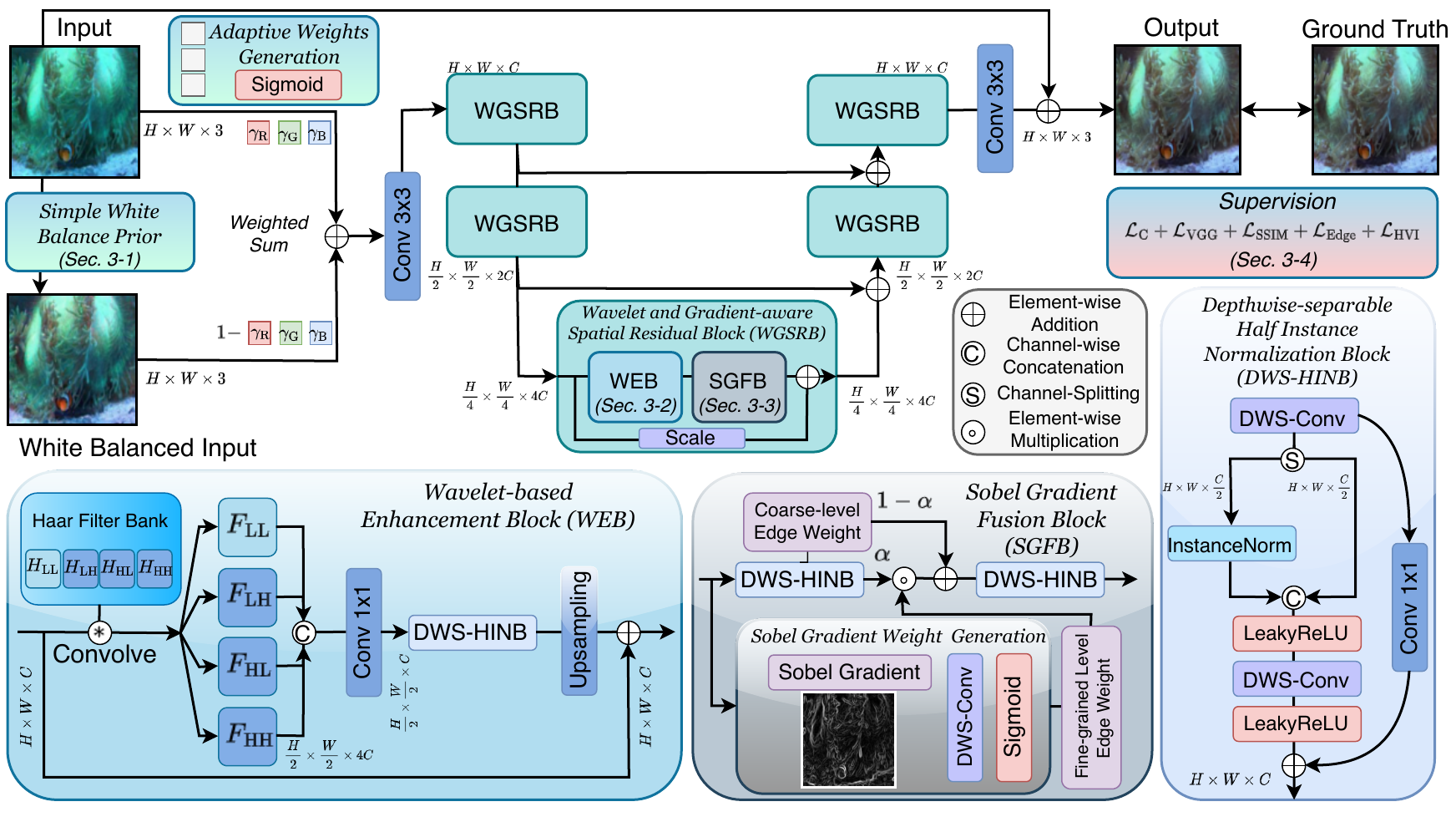}
    \caption{Overall architecture of WWE-UIE. The pipeline applies white balance correction, followed by a U-Net backbone with WGSRBs. Each block integrates a WEB for multi-band decomposition and an SGFB for edge refinement, enabling joint restoration of structure, detail, and color with lightweight efficiency.}
    \label{fig:uie-arch}
\end{figure*}

As illustrated in Fig.~\ref{fig:uie-arch}, the proposed WWE-UIE framework begins with an adaptive white balance module to correct color imbalance. The corrected image is then processed by a U-Net backbone composed of Wavelet and Gradient-aware Spatial Residual Blocks (WGSRBs). Each WGSRB integrates a WEB for efficient wavelet sub-bands feature extraction and an SGFB for edge-preserving refinement. The entire network is trained end-to-end using a task-aligned composite loss for efficient underwater image enhancement.



\begin{algorithm}[ht]
\caption{Simple White Balance}
\label{alg:wb}
\KwIn{Image $X \in \mathbb{R}^{H \times W \times C}$}
\KwOut{White-balanced image $X_{\text{wb}}$}

\BlankLine
$\mu_c \leftarrow \text{mean}_{H,W}(X)$ \tcp*{Per-channel mean}  
$\mu_g \leftarrow \text{mean}_C(\mu_c)$ \tcp*{Gray-world mean}  
$g \leftarrow \mu_g / (\mu_c + \epsilon)$ \tcp*{Channel-wise gain}  
$X \leftarrow X \cdot g$  
$X_{\log} \leftarrow \log(X + \epsilon)$ \;
$X_{\log} \leftarrow X_{\log} - \text{mean}_{H,W}(X_{\log})$ \;
$X_{\text{wb}} \leftarrow \exp(X_{\log})$
$X_{\min} \leftarrow \min(X_{\text{wb}}),\quad
X_{\max} \leftarrow \max(X_{\text{wb}})$ \;
$X_{\text{wb}} \leftarrow \dfrac{X_{\text{wb}} - X_{\min}}{X_{\max} - X_{\min} + \epsilon}$

\Return $X_{\text{wb}}$
\end{algorithm}

\subsection{White Balance Prior}

Underwater images often suffer from severe color imbalance due to wavelength-dependent attenuation, where red light is absorbed much faster than blue and green, resulting in strong blue-green dominance. Such distortions draw the importance of incorporating effective priors into enhancement models: they provide physically meaningful cues that help networks handle the unique degradations of underwater environments.

Existing prior-based approaches, however, have notable limitations. Ucolor \cite{ucolor} employs a reverse medium transmission map derived from hand-crafted transmission estimation, which can guide the network toward degraded regions but remains fragile and unreliable under complex lighting or scattering. More recently, ReXNet \cite{Rex-Net} generates an additional reflectance-based input through explicit reflectance estimation. While this strategy introduces complementary cues, the reflectance estimation step itself is ill-posed, prone to error propagation, and significantly increases model complexity.

To overcome these limitations, we design a robust adaptive white balance prior inspired by the Gray-World assumption \cite{grayworld}, which has proven effective in low-light and tone-mapping tasks \cite{wb-lowlight-tone}. Instead of introducing a new estimation branch or replacing the input entirely, we apply a learnable fusion of the original image $X$ and its white-balanced counterpart $X_{\text{wb}}$:
\begin{equation}
X_{\text{fused}} = \gamma \cdot X_{\text{wb}} + (1 - \gamma) \cdot X,
\end{equation}
where $\gamma = \{\gamma_R, \gamma_G, \gamma_B\}$ are channel-wise learnable parameters constrained to $[0,1]$. This design adaptively regulates the per-channel contribution of the white-balance prior while retaining raw image cues. It directly addresses wavelength-dependent color imbalance without unstable intermediate estimations, enhancing robustness across diverse conditions with minimal computational overhead.

\subsection{Wavelet-based Enhancement Block}

Wavelet decomposition has been widely adopted in image restoration~\cite{wavelet-fuse, wavelet-cnn}, as it jointly preserves spatial and frequency information. Unlike Fourier transforms that provide only global frequency analysis and lose spatial localization, the Discrete Wavelet Transform (DWT) enables multi-band decomposition into low- and high-frequency subbands. This property is particularly beneficial for UIE, where color attenuation mainly affects low-frequency structures and detail loss occurs in high-frequency textures. By explicitly separating these components, DWT allows targeted enhancement of degraded regions while maintaining global structural consistency.

We propose a lightweight WEB that leverages the wavelet transform with a Haar filter bank for efficient multi-band decomposition. The transform is realized via four fixed \(2 \times 2\) convolution kernels. Let $h = \tfrac{1}{2}[1,\,1]$ and $g = \tfrac{1}{2}[1,\,-1]$. The four Haar filters can be compactly expressed as outer products: \begin{equation} H_{\text{LL}} = h^\top h, \quad H_{\text{LH}} = g^\top h, \quad H_{\text{HL}} = h^\top g, \quad H_{\text{HH}} = g^\top g. \end{equation} These kernels decompose the input into one approximation component (LL) and three directional detail components (LH: horizontal, HL: vertical, HH: diagonal). For an input feature map \(F_{\text{in}} \in \mathbb{R}^{H \times W \times C}\), the four subbands are obtained as: \begin{equation} \{F_{\text{LL}}, F_{\text{LH}}, F_{\text{HL}}, F_{\text{HH}}\} = \text{DWT}(F_{\text{in}}). \end{equation} 

\begin{figure}[H]
    \centering
    \includegraphics[width=1.0\linewidth]{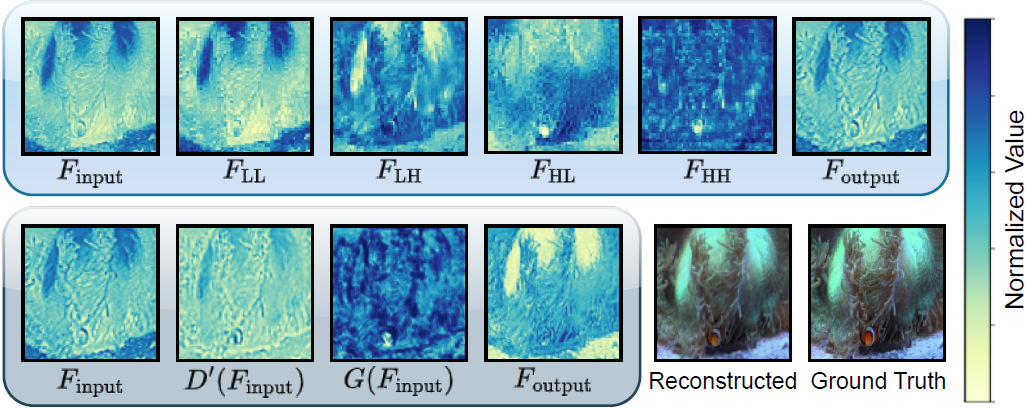}
    \caption{Visualization of feature maps within WEB and SGFB. The WEB highlights multi-band decomposition into structural and textural components, while the SGFB emphasizes gradient-based enhancement of edges and fine details.
    }

    \label{fig:feature-map}
\end{figure}

As illustrated in Fig.~\ref{fig:feature-map}, the four subbands are concatenated and compressed by a \(1 \times 1\) convolution:
\begin{equation}
    F_{\text{fused}} = \text{Conv}_{1\times1}(\text{Concat}(F_{\text{LL}}, F_{\text{LH}}, F_{\text{HL}}, F_{\text{HH}})).
\end{equation}

The fused features are then refined by the DWS-HINB module~\cite{dws-cnn,HIN}, denoted as $D(\cdot)$, which adapts to instance-specific color shifts in underwater images. The final output is formulated as:
\begin{equation}
    F_{\text{out}} = \text{Upsample}(D(F_{\text{fused}})) + F_{\text{in}}.
\end{equation}

Unlike WF-Diff~\cite{wf-diff}, which integrate frequency cues with the diffusion model at high computational cost, our design achieves efficient frequency-aware enhancement with lightweight fixed filters. TCTL-Net~\cite{TCTL}, while leveraging color transfer and self-attention, overlooks explicit frequency decomposition, limiting its ability to restore fine structures. In contrast, WEB explicitly separates subbands, enabling both global consistency and local detail recovery with minimal overhead.


\subsection{Sobel Gradient Fusion Block}

In addition to color distortion, one of the main challenges in UIE is scattering, which causes optical blurring due to suspended particles in water. This results in weakened edges and loss of structural clarity, making edge-aware enhancement crucial for detail restoration.

To address this, we propose the SGFB, which leverages gradient priors for fine-grained feature refinement. Prior works have shown that gradient guidance improves restoration quality by strengthening structural cues. As illustrated in Fig.~\ref{fig:feature-map}, the input feature map is first refined by a DWS-HINB module, denoted as \(D'(\cdot)\):
\begin{equation}
    F_{0} = D'(F_{\text{in}}).
\end{equation}

In parallel, we adopt Sobel operators, which provide robust first-order directional gradients. Unlike Laplacian filters that are noise-sensitive and prone to amplifying particle-induced fluctuations, Sobel emphasizes edge orientation and just perfectly aligns with wavelet-decomposed subbands, making it better suited for underwater images where scattering blurs edges but preserves directional cues. Let
\begin{equation}
    G(F_{0}) = \sqrt{(F_{0} * S_x)^2 + (F_{0} * S_y)^2},
    \label{eq:sobel}
\end{equation}
where \(S_x\) and \(S_y\) are the horizontal and vertical Sobel kernels, and \(*\) denotes convolution. 
The gradient features are then compressed by a depthwise-separable convolution with sigmoid activation to produce a gating map:
\begin{equation}
    M = \sigma(\text{DWS-Conv}_{3\times3}(G(F_{0}))).
\end{equation}

This gating map \(M\) provides fine-grained, pixel-level modulation by highlighting edge regions, while the learnable scalar \(\alpha \in [0,1]\) serves as a coarse-level controller that balances the contribution of gradient-enhanced and baseline features. The fused representation is expressed as
\begin{equation}
    F_{1} = \alpha \cdot (M \odot F_{0}) + (1 - \alpha) \cdot F_{0},
\end{equation}
where \(\odot\) denotes element-wise multiplication. Finally, a second DWS-HINB module \(D''(\cdot)\) further refines the fused features:
\begin{equation}
    F_{\text{out}} = D''(F_{1}),
\end{equation}
producing an edge-preserving output that strengthens structural boundaries while keeping homogeneous regions stable for underwater image enhancement.

Compared with ReX-Net \cite{Rex-Net}, which integrates spatial–channel attention to yield weighted feature maps in an implicit manner, SGFB provides explicit gradient-aware regulation. SFGNet~\cite{sfgnet} also leverages gradient cues, but its gating depends on multiple convolutional layers with additional parameters, making it more complex and less transparent. In contrast, our design combines Sobel-derived gradient maps for fine-grained modulation with a lightweight scalar gate for coarse-level control, delivering clearer edge enhancement with minimal overhead and achieving sharper, more stable restoration under scattering.

\subsection{Loss Function}
Effective supervision of UIE requires a carefully designed objective function. Consequently, we employ a composite loss that addresses the heterogeneous degradations characteristic of underwater imagery. Assumed that $Y$ is the reference image and $Y'$ the restored result, we define the losses as follows:

\textbf{Charbonnier Loss}.  
A smooth variant of the L1 loss that stabilizes training:
\begin{equation}
\mathcal{L}_{\text{C}}(Y, Y') = \sqrt{(Y - Y')^2 + \epsilon^2},
\end{equation}
where $\epsilon$ is a negligible constant to avoid divergence.

\textbf{SSIM Loss}.  
Encourages structural similarity in terms of luminance, contrast, and texture:
\begin{equation}
\mathcal{L}_{\text{SSIM}}(Y, Y') = 1 - \text{SSIM}(Y, Y').
\end{equation}

\textbf{Perceptual Loss}.  
Maintains semantic and perceptual realism via VGG feature distance:
\begin{equation}
\mathcal{L}_{\text{VGG}}(Y, Y') = \tfrac{1}{N} \sum_{i=1}^{N} \left\|\phi^{(i)}(Y) - \phi^{(i)}(Y')\right\|_2,
\end{equation}
where $\phi^{(i)}(\cdot)$ are features from the $i$-th VGG16 layer.

\textbf{Edge-aware Loss}.  
Preserves boundaries using Sobel gradient differences:
\begin{equation}
\mathcal{L}_{\text{Edge}}(Y, Y') = \left\| G(Y) - G(Y') \right\|_2,
\end{equation}
where $G(\cdot)$ is defined in Eq.~\ref{eq:sobel}.

\textbf{HVI Loss}.  
Improves color fidelity in a more stable space. Unlike HSV, which suffers from a red-hue discontinuity, or CIELab, which exhibits nonlinear hue distortions~\cite{labnotlin}, HVI decouples chromaticity and intensity for robust supervision:
\begin{equation}
\mathcal{L}_{\text{HVI}}(Y, Y') = \mathcal{L}_{\text{C}}\left( \text{HVI}(Y), \text{HVI}(Y') \right).
\end{equation}
This yields smoother gradients in back-prop and more stable color supervision under severe casts.

\textbf{Total Loss}.  
The overall objective is a weighted sum:
\begin{equation}
\mathcal{L}_\text{total} = \lambda_{1}\mathcal{L}_\text{C} + \lambda_{2}\mathcal{L}_\text{VGG} + \lambda_{3}\mathcal{L}_\text{SSIM} + \lambda_{4}\mathcal{L}_\text{Edge} + \lambda_{5}\mathcal{L}_\text{HVI},
\end{equation}
with $\lambda_1=1$, $\lambda_2=0.1$, $\lambda_3=0.1$, $\lambda_4=0.4$, and $\lambda_5=0.5$ in all experiments.

\begin{table*}[h]
\centering
\caption{Performance and complexity comparison on \textbf{full-reference datasets}.}

\label{tab:psnr_ssim_params_flops}
\renewcommand\arraystretch{1.3}
\tabcolsep=0.18cm
\scalebox{0.67}{
\begin{tabular}{l|cc|cc|cc|cc|cc|c|c|c}
\toprule[0.15em]
\multirow{2}{*}{\textbf{Methods}} 
 & \multicolumn{2}{c|}{\textbf{UIEB}} 
 & \multicolumn{2}{c|}{\textbf{LSUI}} 
 & \multicolumn{2}{c|}{\textbf{UFO-120} }
 & \multicolumn{2}{c|}{\textbf{EUVP-Scene}} 
 & \multicolumn{2}{c|}{\textbf{EUVP-Dark}} 
 & \multirow{2}{*}{\textbf{Params (M)}} 
 & \multirow{2}{*}{\textbf{FLOPs (G)}} 
 & \multirow{2}{*}{\textbf{Runtime (ms)}} \\
\cline{2-11}
 & PSNR↑ & SSIM↑ & PSNR↑ & SSIM↑ & PSNR↑ & SSIM↑ & PSNR↑ & SSIM↑ & PSNR↑ & SSIM↑ &  &  &  \\
\hline
NU2Net (\textit{AAAI'23})  & 22.9641 & 0.8951 & 25.3280 & 0.8887 & 16.2720 & 0.6742 & 24.2798 & 0.7861 & 21.8420 & 0.8906 & 3.146  & 10.486 & \cellcolor{red!25}{\textbf{2.291}} \\
UShape (\textit{TIP'23}) & 21.5493 & 0.8064 & 27.5431 & 0.8734 & 26.6535 & 0.8201 & 24.9338 & 0.7463 & 21.6864 & 0.8763 & 22.817 &  \cellcolor{orange!25}{\textbf{2.983}} & 41.764 \\
UDAFormer (\textit{C\&G'23}) & 23.5020 & 0.8955 & \cellcolor{yellow!25}{\textbf{30.7702}} & 0.9236 & \cellcolor{red!25}{\textbf{28.5539}} & \cellcolor{orange!25}{\textbf{0.8605}} & \cellcolor{red!25}{\textbf{26.6110}} & 0.8118 & 22.1975 & 0.8931 & 9.594  & 41.593 & 80.630 \\
LiteEnhanceNet (\textit{ESWA'24})   & 22.3930 & 0.9009 & 25.3363 & 0.9048 & 26.0682 & 0.8481 & 25.0688 & 0.8044 & 21.7822 & 0.9015 & \cellcolor{red!25}{\textbf{0.013}}  & \cellcolor{red!25}{\textbf{0.690}} & \cellcolor{orange!25}{\textbf{2.356}} \\
SFGNet (\textit{ICASSP'24}) & 21.0876 & 0.8777 & 24.9309 & 0.9010 & 25.4102 & 0.8552 & 25.7528 & \cellcolor{orange!25}{\textbf{0.8192}} & 22.0395 & 0.9039 & 1.298  & 81.575 & 16.723 \\
SMDR–IS (\textit{AAAI'24})  & 23.4257 & \cellcolor{orange!25}{\textbf{0.9119}} & 30.4474 & \cellcolor{orange!25}{\textbf{0.9298}} & \cellcolor{yellow!25}{\textbf{28.2575}} & \cellcolor{yellow!25}{\textbf{0.8604}} & \cellcolor{yellow!25}{\textbf{26.5022}} & 0.8087 & \cellcolor{orange!25}{\textbf{22.3854}} & \cellcolor{red!25}{\textbf{0.9085}} & 12.253 & 48.318 & 36.586 \\
CDF–UIE (\textit{TGRS'25})  & 22.3103 & 0.8635 & 29.5376 & 0.9188 & 27.7569 & 0.8552 & 26.3730 & 0.8082 & 22.0485 & 0.8956 & 15.902 & 57.457 & 24.404 \\
PhaseFormer (\textit{WACV'25})  & \cellcolor{yellow!25}{\textbf{23.7561}} & 0.9084 & \cellcolor{red!25}{\textbf{32.1155}} & 0.9232 & 27.4412 & 0.8069 & 25.9013 & 0.7630 & 22.1223 & 0.8951 & 1.779  & 13.041 & 40.657 \\
\hline
\textbf{WWE-UIE (Ours-Efficient)}   & \cellcolor{orange!25}{\textbf{23.9176}} & \cellcolor{yellow!25}{\textbf{0.9109}} & 30.2849 & \cellcolor{yellow!25}{\textbf{0.9277}} & 27.8165 & 0.8599 & 26.4562 & \cellcolor{yellow!25}{\textbf{0.8185}} & \cellcolor{yellow!25}{\textbf{22.3452}} & \cellcolor{yellow!25}{\textbf{0.9078}} & \cellcolor{orange!25}{\textbf{0.471}}   &  \cellcolor{yellow!25}{\textbf{3.603}} & \cellcolor{yellow!25}{\textbf{7.111}} \\
\textbf{WWE-UIE (Ours)}  & \cellcolor{red!25} \textbf{24.3209} & \cellcolor{red!25}\textbf{0.9196} & \cellcolor{orange!25}{\textbf{31.0238}} & \cellcolor{red!25}{\textbf{0.9310}} & \cellcolor{orange!25}{\textbf{28.2840}} & \cellcolor{red!25}{\textbf{0.8668}} & \cellcolor{orange!25}{\textbf{26.5923}} & \cellcolor{red!25}{\textbf{0.8243}} &  \cellcolor{red!25}{\textbf{22.5327}} & \cellcolor{orange!25}{\textbf{0.9083}} & \cellcolor{yellow!25}{\textbf{0.734}} & 6.251 & 7.602 \\
\bottomrule[0.15em]
\end{tabular}
}
\end{table*}

\begin{figure*}[t]
    \centering
    \includegraphics[width=1.0\linewidth]{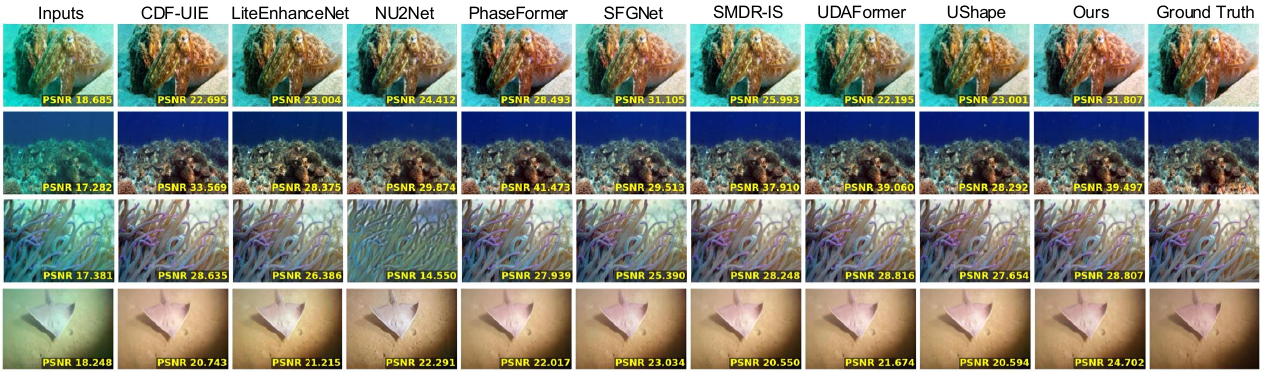}
    
    \caption{Visual comparison on \textbf{full‐reference} datasets. Each row corresponds to the datasets \textbf{\textit{UIEB}}, \textbf{\textit{LSUI}}, \textbf{\textit{UFO}}, and \textbf{\textit{EUVP-Scene}}.}

    \label{fig:full-ref-plot}
\end{figure*}

\section{Experiments}

\subsection{Experiment Settings}
\label{sec:exp-settings}

\textbf{Compared Methods}. 
We compare WWE-UIE with nine recent state-of-the-art UIE methods to contextualize its performance. 
The compared methods cover three categories:  
(i) \textit{Spatial–frequency integrated models}: SFGNet~\cite{sfgnet}, CDF-UIE~\cite{cdfuie}, and PhaseFormer~\cite{phaseformer}, which explicitly couple spatial and frequency cues for restoration;  
(ii) \textit{High-capacity designs}: UShape~\cite{ushape}, UDAFormer~\cite{udaformer}, and SMDR-IS~\cite{smdris}, which employ transformer or diffusion backbones to capture global dependencies but incur heavy computational costs; 
(iii) \textit{Lightweight networks}: NU2Net~\cite{uranker} and LiteEnhance~\cite{liteenhancenet}, designed with efficiency as the primary objective.

\textbf{Benchmark Datasets.} Since paired ground truth is often unavailable in real underwater scenes, evaluation for UIE is conducted on both \textit{full-reference synthetic datasets} (for quantitative fidelity) and \textit{non-reference real-world datasets} (for practical robustness). (1) \textit{Full-reference datasets}: We train our model on five benchmark synthetic datasets: UIEB~\cite{waternet}, UFO-120~\cite{ufo120}, LSUI~\cite{ushape}, and EUVP~\cite{FUnIE-GAN} (including EUVP-Dark and EUVP-Scene). These datasets contain 890, 1620, 4279, 2185, and 5500 paired images and ground truth, respectively.

(2) \textit{Non-reference datasets}: We evaluate the UIEB-pretrained model on Challenging-60~\cite{waternet} and U45~\cite{u45}, and further assess color correction performance on the Color-Check7~\cite{cc7} dataset. The Challenging-60 dataset is a real-world subset from UIEB, specifically curated for difficult scenes, while U45 comprises various real-world underwater images with diverse degradation conditions. Color-Check7 comprises seven images captured by different camera models for evaluating color fidelity.

\begin{figure}[H]
    \centering
    \includegraphics[width=0.95\linewidth]{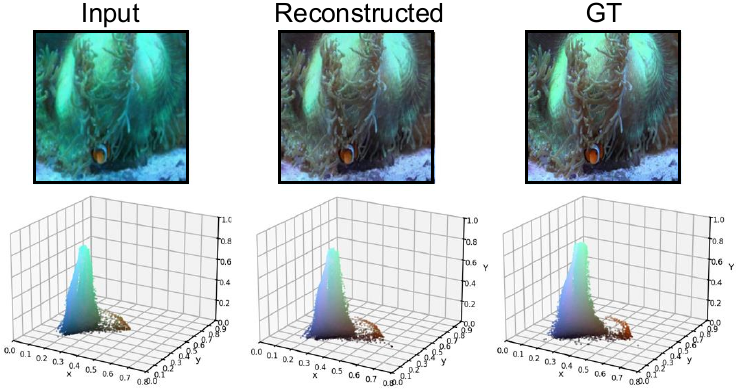}
    \caption{Visualization of our results in the CIE~xyY color space, showing improved alignment of chromaticity and luminance distributions with the reference.}

    \label{fig:xyY}
\end{figure}

\begin{figure*}
    \centering
    \includegraphics[width=1.0\linewidth]{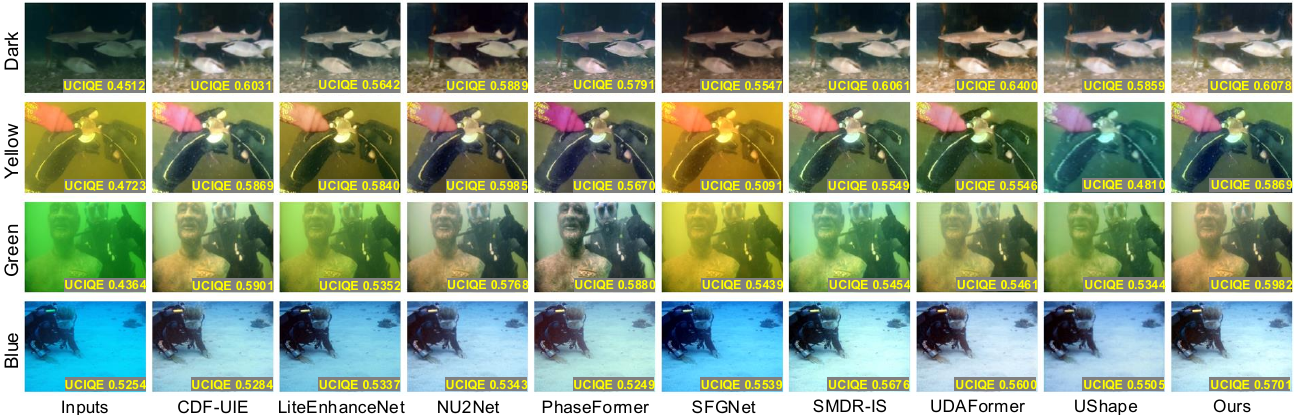}
    \caption{Visual comparison on \textbf{non‐reference} datasets under four different extreme color‐bias conditions, with annotated UCIQE scores. The top two rows correspond to \textbf{\textit{Challenging-60}}, and the bottom two rows correspond to \textbf{\textit{U45}}.}
    \label{fig:non-ref-plot}
\end{figure*}

\textbf{Evaluation Metrics.} To evaluate restoration quality, we adopt PSNR and SSIM on full-reference datasets. PSNR measures pixel-wise fidelity, while SSIM captures structural similarity aligned with human perception. For non‐reference datasets, we use UCIQE~\cite{uciqe}, NIQE~\cite{niqe}, and the learning‐based URanker~\cite{uranker} to assess colorfulness, contrast, and perceptual quality. We favor NIQE over the common UIQM~\cite{uiqm} because UIQM shows lower rank‐order correlation with human judgments~\cite{uranker}, and NIQE’s focus on natural images better reflects the general goal of removing underwater color cast toward terrestrial appearance. Additionally, CIEDE2000~\cite{ciede2000} is employed to quantify color correction accuracy by measuring perceptual color differences on the Color-Check7 dataset \cite{cc7}.

\textbf{Implementation Details.} We implement our method in PyTorch and run all experiments on a single NVIDIA RTX 3090. The inputs are resized to $256\times256$ during training. We use AdamW~\cite{adamw} with an initial learning rate of $2\times10^{-3}$, which is annealed with cosine decay to $1\times10^{-4}$ of the initial value. Each model is trained for 100 epochs with a batch size of 24. Each dataset is randomly split into training, validation, and testing subsets with a ratio of 8:1:1 as default experiment settings. The channel dimension $C$ is set to 32. We employ the standard augmentation with random rotations and flips during training.

\subsection{Results Analysis on Synthetic Datasets}
\label{sec:ref-analy}

\textbf{Quantitative Results.}
As shown in Tab.~\ref{tab:psnr_ssim_params_flops}, our method consistently achieves the highest or second-highest PSNR/SSIM scores. Notably, WWE-UIE delivers both higher restoration quality and faster inference than most competing methods, outperforming all but the explicitly lightweight LiteEnhanceNet \cite{liteenhancenet}, which trades efficiency for noticeably lower quantitative results.

\textbf{Model Efficiency.}
We also provide a reduced variant (\textbf{Ours-E}) by lowering the embedding dimension. This variant further decreases parameters and FLOPs with only a marginal performance drop, highlighting the advantage of our prior-integrated architecture. However, FLOPs alone cannot fully reflect inference efficiency. To this end, we measured the average inference time over 1,000 runs on RGB images of size $256 \times 256$. While LiteEnhanceNet \cite{liteenhancenet} achieves a slightly faster runtime, its quality is significantly lower. WWE-UIE achieves the balance of accuracy and efficiency among state-of-the-art methods.

\textbf{Visual Quality.}
As illustrated in Fig.~\ref{fig:full-ref-plot}, our model restores natural colors and sharper structures compared to state-of-the-art methods. These results validate its effectiveness in addressing diverse underwater degradations while maintaining computational efficiency.

\textbf{Color Distribution.}
To further analyze color fidelity, we visualize restored outputs in the CIE 1931 xyY color space (Fig.~\ref{fig:xyY}). WWE-UIE’s distribution closely matches the ground truth in both chromaticity and luminance, demonstrating effective correction of color casts and illumination imbalances.

\subsection{Results Analysis on Real-world Dataset}

\textbf{Overall Robustness.}
To assess robustness in practical settings, we evaluate on two challenging real-world underwater datasets: Challenging-60 \cite{waternet} and U45 \cite{u45}. As shown in Tab.~\ref{tab:non_ref_performance}, WWE-UIE achieves competitive performance on both datasets, balancing visual quality and consistency.

\textbf{Visual Comparisons.}
Representative results in Fig.~\ref{fig:non-ref-plot} further verify the effectiveness of our  strategy under complex illumination, scattering, and severe color casts. In four extreme bias cases (dark, yellow, green, blue), our model restores more natural tones and sharper structures than many recent state-of-the-art approaches.

\definecolor{first}{RGB}{255, 0, 0}
\definecolor{second}{RGB}{3, 200, 56}
\definecolor{third}{RGB}{6, 95, 228}
\begin{table}[h]
\centering
\renewcommand\arraystretch{1.15}
\setlength{\tabcolsep}{3pt}
\caption{Quantitative comparison on non-reference datasets. 
}
\resizebox{0.478\textwidth}{!}{
\begin{tabular}{l|ccc|ccc|c}
\toprule[0.1em]
\multirow{2}{*}{\textbf{Methods}} & \multicolumn{3}{c|}{\textbf{Challenging-60}} & \multicolumn{3}{c|}{\textbf{U45}} & \multicolumn{1}{c}{\textbf{Color-Check7}}\\
& UCIQE ↑ & NIQE ↓ & URanker ↑ & UCIQE ↑ & NIQE ↓ & URanker ↑ & CIEDE2000 ↓\\
\hline
Input         & 0.4800 & 8.5538 & -0.0873 & 0.4814 & 4.5470 & -0.2743 & 14.4598 \\
\hline
NU2Net        & 0.5692 & 6.1968 & \cellcolor{yellow!25}{\textbf{1.4212}} & 0.5855 & 4.3935 & \cellcolor{red!25}{\textbf{1.9096}} & 10.9393\\
UShape         & 0.5644 & \cellcolor{orange!25}{\textbf{5.9620}} & 0.4655 & 0.5886 & 4.5788 & 1.8588 & 12.0191\\
UDAFormer     & \cellcolor{red!25}{\textbf{0.5841}} & 6.2011 & \cellcolor{orange!25}{\textbf{1.4335}} & \cellcolor{yellow!25}{\textbf{0.6009}} & 4.3870 & 1.7897 & \cellcolor{yellow!25}{\textbf{10.4507}}\\
LiteEnhanceNet & 0.5596 & 6.1968 & 1.1667 &  0.5880 & 4.3935 & 1.4425 & 12.3064\\
SFGNet         & 0.5485& 6.7104 & 0.9094 & 0.5805 & 5.6445 & 1.2187 & 12.9670\\
SMDR–IS      & \cellcolor{orange!25}{\textbf{0.5819}} & 6.3385 & 1.2925 & \cellcolor{orange!25}{\textbf{0.6060}} & \cellcolor{yellow!25}{\textbf{4.1404}} & \cellcolor{yellow!25}{\textbf{1.8817}} & 11.2903\\
CDF–UIE     & 0.5759 & \cellcolor{yellow!25}{\textbf{6.0359}} & \cellcolor{red!25}{\textbf{1.4501}} & 0.5981 & 4.7315 & 1.8159 & \cellcolor{orange!25}{\textbf{10.3654}}\\
PhaseFormer  & 0.5421 & 6.3261 & 1.1642 & 0.5795 & \cellcolor{red!25}{\textbf{3.9684}} & 1.8628 & 10.8478\\
\hline
\textbf{WWE-UIE (Ours)} & \cellcolor{yellow!25}{\textbf{0.5784}} & \cellcolor{red!25}{\textbf{5.9575}} & 1.4178 & \cellcolor{red!25}{\textbf{0.6110}} & \cellcolor{orange!25}{\textbf{4.1057}} & \cellcolor{orange!25}{\textbf{1.8992}} & \cellcolor{red!25}{\textbf{10.0979}}\\
\bottomrule[0.1em]
\end{tabular}}
\label{tab:non_ref_performance}
\end{table}

\begin{figure}[H]
    \centering
    \includegraphics[width=1.0\linewidth]{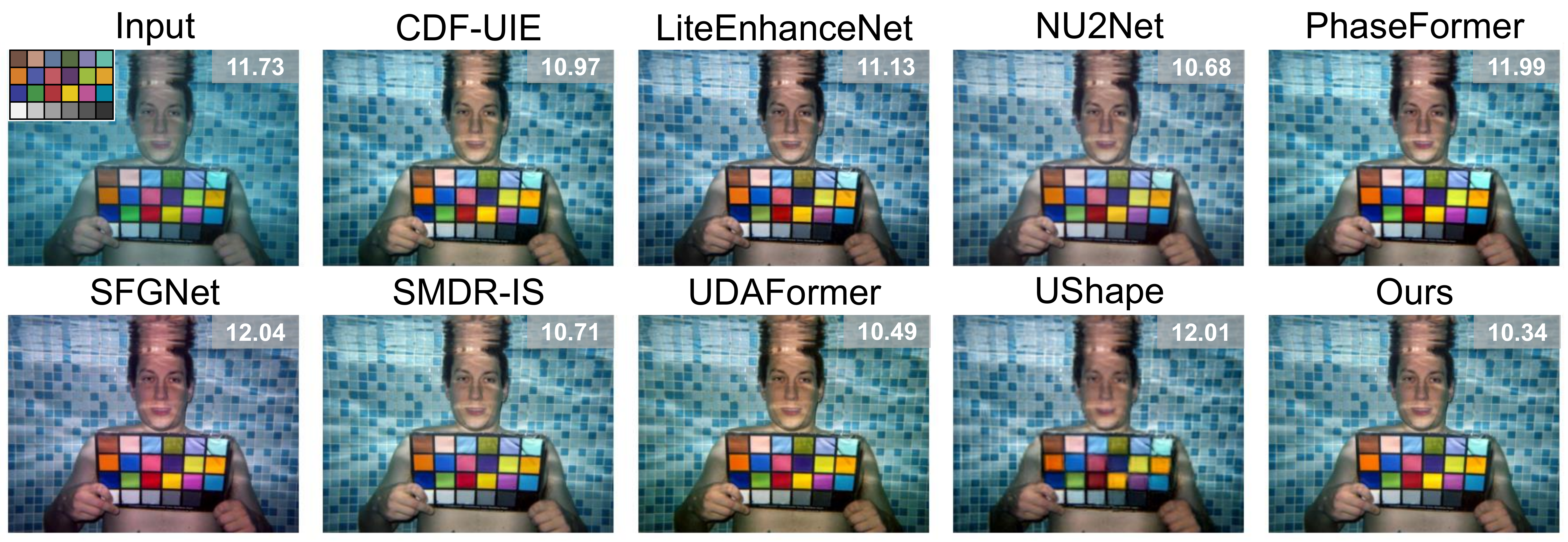}
    \caption{Color restoration on a Color-Check7 image (Oly T8000). \textbf{CIEDE2000} scores are shown at the top right, with the reference chart overlaid on left-upper side of the \textbf{Input}.}
    \label{fig:cc7}
\end{figure}

\textbf{Comparative Analysis.}
While CDF–UIE \cite{cdfuie} and PhaseFormer \cite{phaseformer} achieve the top URanker on Challenging-60~\cite{waternet} and the best NIQE on U45 \cite{u45}, respectively, their performance varies across datasets. In contrast, WWE-UIE delivers more consistent results overall, indicating stronger generalization. Competing lightweight methods such as LiteEnhanceNet \cite{liteenhancenet} or SFGNet \cite{sfgnet} often leave residual haze and color distortion, and PhaseFormer \cite{phaseformer} occasionally produces bluish over-enhancement.

\textbf{Color Correction.}
We further evaluate color fidelity using the Color-Check7 dataset \cite{cc7} with a model pretrained on UIEB. As reported in Tab.~\ref{tab:non_ref_performance}, WWE-UIE achieves superior CIEDE2000 scores, confirming its advantage in restoring perceptually accurate colors. A representative case is shown in Fig.~\ref{fig:cc7}, where our output aligns closely with the reference chart.

\begin{table}[ht]
\centering
\caption{Ablation studies on \textbf{loss functions}.}
\label{tab:abl-loss}
\begin{threeparttable}
\resizebox{0.95\linewidth}{!}{
\begin{tabular}{c|c c c c|cc}
\toprule \textbf{Configuration}
 & $\mathcal{L}_\text{SSIM}$ & $\mathcal{L}_\text{VGG}$ & $\mathcal{L}_\text{Edge}$ & $\mathcal{L}_\text{HVI}$ & {PSNR ↑} & {SSIM ↑} \\
\hline
Baseline (Only $\mathcal{L}_\text{C}$)
& \ding{55} & \ding{55} & \ding{55} & \ding{55} & 24.0614 & 0.9064 \\
w/o $\mathcal{L}_\text{SSIM}$ & \ding{55} & \ding{51} & \ding{51} & \ding{51} & 24.1462 & 0.9138 \\
w/o $\mathcal{L}_\text{VGG}$  & \ding{51} & \ding{55} & \ding{51} & \ding{51} & 24.1154 & 0.9130 \\
w/o $\mathcal{L}_\text{Edge}$ & \ding{51} & \ding{51} & \ding{55} & \ding{51} & 24.3034 & 0.9144 \\
w/o $\mathcal{L}_\text{HVI}$  & \ding{51} & \ding{51} & \ding{51} & \ding{55} & 23.8852 & 0.9131 \\
\hline
\textbf{Full Model}      & \ding{51} & \ding{51} & \ding{51} & \ding{51} & \textbf{24.3209} & \textbf{0.9196} \\
\bottomrule
\end{tabular}
}

\end{threeparttable}
\end{table}

\subsection{Ablation Study}
To investigate the contribution of each component in our loss design and architecture, we conduct a series of ablation experiments on the UIEB dataset. 

\begin{figure}[H]
    \centering
    \includegraphics[width=1.01\linewidth]{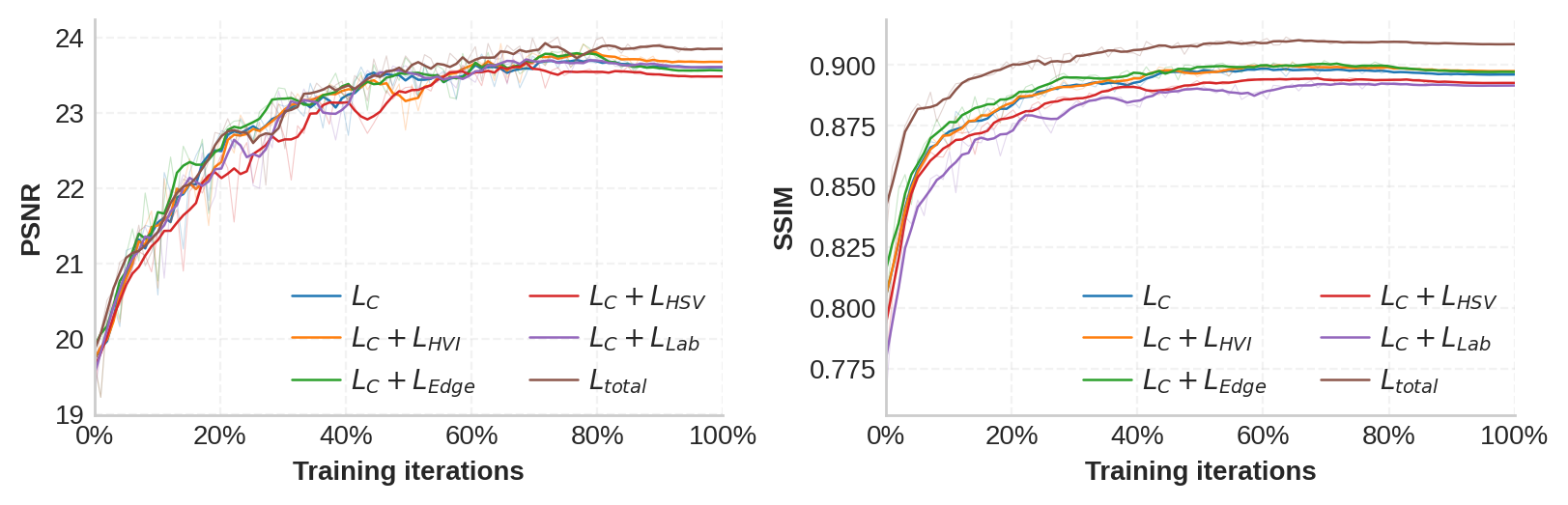}
    \caption{Optimization curves for $\mathcal{L}_{\text{edge}}$ and $\mathcal{L}_{\text{HVI}}$. Combining both yields the best results and faster convergence speed.}
    \label{fig:curve}
\end{figure}

\textbf{Loss Function Analysis}.
As shown in Tab.~\ref{tab:abl-loss}, we investigate the effects of various loss term combinations. Beginning with the basic  loss $\mathcal{L}_\text{C}$, the progressive inclusion of $\mathcal{L}_{\text{SSIM}}$, $\mathcal{L}_{\text{VGG}}$,  $\mathcal{L}_{\text{Edge}}$, and $\mathcal{L}_{\text{HVI}}$ leads to consistent performance improvements. Notably, removing all auxiliary losses significantly reduces SSIM, while excluding the $\mathcal{L}_{\text{HVI}}$ results in the lowest PSNR. These results demonstrate the complementary roles of each loss component in enhancing structural fidelity, perceptual quality, and color restoration. Fig.~\ref{fig:curve} further shows that combining $\mathcal{L}_{\text{Edge}}$ and $\mathcal{L}_{\text{HVI}}$ yields faster convergence, whereas $\mathcal{L}_{\text{HSV}}$ and $\mathcal{L}_{\text{CIELab}}$ are insufficient and even degrade performance compared with using the $\mathcal{L}_\text{C}$ alone.

\begin{table}[ht]
    \centering
    \caption{Ablation studies on \textbf{architectural modules}.}
    \label{tab:abl-block}
    \resizebox{0.89\linewidth}{!}{
    \begin{tabular}{l|cc}
    \toprule
    \textbf{Configuration} & {PSNR ↑} & {SSIM ↑} \\
    \hline
    w/o white-balance prior                 & 24.1820 & 0.9175 \\
    w/o WEB                                 & 24.2210 & 0.9134 \\
    w/o gradient-aware branch in SGFB       & 23.9047 & 0.9143 \\
    Order (WGSRB): $\text{SGFB}\!\to\!\text{WEB}$ (swapped) & 24.0622 & 0.9154 \\
    \hline
    \textbf{Full Model}                    & \textbf{24.3209} & \textbf{0.9196} \\
    \bottomrule
    \end{tabular}%
    }
\end{table}

\textbf{Module Contribution}.
We further assess the contribution of each proposed module in our architecture. As presented in Tab.~\ref{tab:abl-block}, removing the white balance prior, the WEB, or the gradient-aware mechanism in SGFB results in performance degradation. Specifically, the white balance prior contributes more to structural similarity, while the gradient-aware design enhances pixel-wise accuracy. Furthermore, we analyze the module order within WGSRB and find that applying WEB first to extract multi-band features, followed by SGFB to refine boundary textures, is better suited for UIE tasks. In summary, the full model achieves the strongest overall performance, supporting the effectiveness of our integrated design.

\begin{figure}[H]
    \centering
    \includegraphics[width=1.0\linewidth]{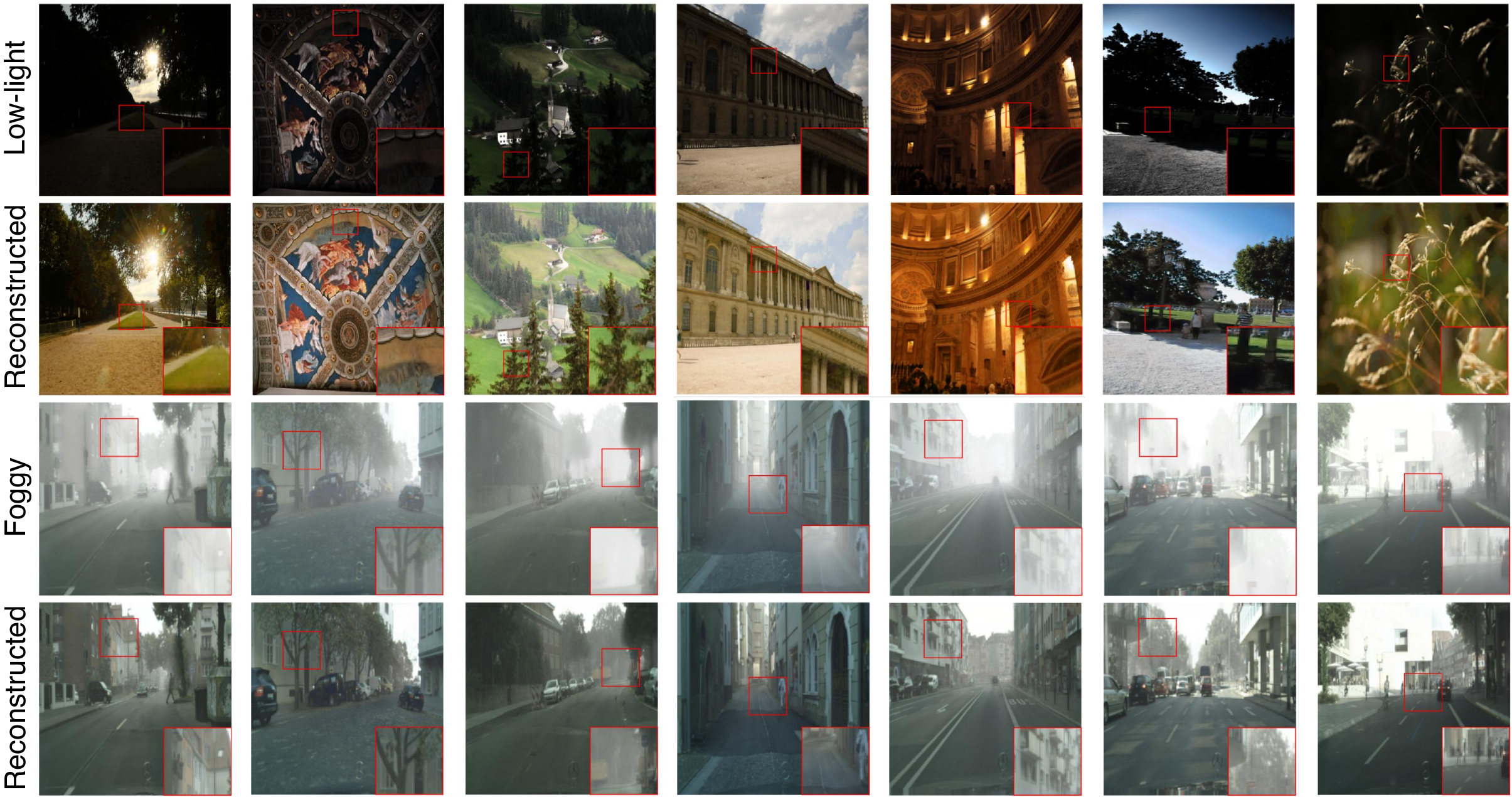}
    \caption{Generalization results of WWE-UIE on LOL-v2 (low-light) \cite{lolv2} and Foggy Cityscapes (dehazing) \cite{foggycity}. Our model shows desired results beyond underwater restoration, with clearer textures and improved contrast. (Zoom-in for details).}

\label{fig:low-light}
\end{figure}

\textbf{Cross-domain Restoration Capability}.
While WWE-UIE is primarily designed for underwater enhancement, we also conducted exploratory experiments on other degradation types. As shown in Fig.~\ref{fig:low-light}, the model demonstrates promising potential in handling low-light and foggy scenes, suggesting that the prior-guided design may extend beyond UIE to broader restoration tasks.

\section{Conclusions}
In this paper, we introduced \textbf{WWE-UIE}, a novel UIE framework that unifies adaptive white balance, wavelet-based multi-band decomposition, and edge-aware refinement within a single architecture. First, WEB decomposes feature maps into wavelet subbands, thereby recovering global structures and local fine textures. Second, SGFB explicitly regulates edge regions with dual-level weight maps to preserve sharp boundaries under scattering. Extensive experiments on both reference and non-reference benchmarks show that our method achieves competitive restoration quality with substantially fewer parameters and faster inference.

\bibliographystyle{ieeenat_fullname}
\bibliography{main}

\end{document}